\documentclass{ISMA_USD}

\hypersetup{pdftitle  = {Towards an active-learning approach to resource allocation for population-based damage prognosis},
	pdfauthor = {G. Tsialiamanis ,K. Worden, N. Dervilis, A.J. Hughes},
	pdfkeywords = {Population-based structural health monitoring (PBSHM), damage prognosis, machine learning, active learning, resource allocation}}

\title{Towards an active-learning approach to resource allocation for population-based damage prognosis}

\author[1]{G.\ Tsialiamanis}
\author[1]{K.\ Worden}
\author[1]{N.\ Dervilis}
\author[1]{A.J.\ Hughes}

\affil[1]	{University of Sheffield, Sheffield, Department of Mechanical Engineering,\NewLineAffil
			Mappin St, Sheffield, United Kingdom \NewAffil}
		
\date{}

\begin{document}



\abstract{Damage prognosis is, arguably, one of the most difficult tasks of structural health monitoring (SHM). To address common problems of damage prognosis, a population-based SHM (PBSHM) approach is adopted in the current work. In this approach the prognosis problem is considered as an information-sharing problem where data from past structures are exploited to make more accurate inferences regarding currently-degrading structures. For a given population, there may exist restrictions on the resources available to conduct monitoring; thus, the current work studies the problem of allocating such resources within a population of degrading structures with a view to maximising the damage-prognosis accuracy. The challenges of the current framework are mainly associated with the inference of outliers on the level of damage evolution, given partial data from the damage-evolution phenomenon. The current approach considers an initial population of structures for which damage evolution is extensively observed. Subsequently, a second population of structures with evolving damage is considered for which two monitoring systems are available, a low-availability and high-fidelity (low-uncertainty) one, and a widely-available and low-fidelity (high-uncertainty) one. The task of the current work is to follow an active-learning approach to identify the structures to which the high-fidelity system should be assigned in order to enhance the predictive capabilities of the machine-learning model throughout the population.}

\maketitle


\section{Introduction}
\label{sec:intro}

The need to maintain the integrity and safety of structures has led to the development of systems to monitor their health state. The field of structural dynamics which deals with the identification of structural damage is that of \textit{structural health monitoring} (SHM) \cite{Farrar}. The procedure of identifying damage in structures may often be dealt with using engineering intuition; however it is most commonly addressed via \textit{data-driven} techniques. Naturally, monitoring shall require even a minimum level of data acquisition, in order to characterise the actual state of a structure. Such data are also exploited for the development of machine-learning \cite{Bishop2, goodfellow2016deep} models which learn to perform the various tasks. Machine-learning models offer the convenience of defining models directly using data and bypassing the need to mathematically formulate a physics-based model, which, especially given the complexity of modern structures and materials, may be quite tedious or even impossible.

The several tasks of SHM are summarised by Rytter's hierarchy \cite{rytter1993vibrational} which was later extended in \cite{Farrar}. The tasks refer to i) the \textit{existence} of damage, ii) the \textit{location} of damage, iii) the \textit{classification} of damage, iv) the \textit{extent} of damage, and v) the \textit{prognosis} of damage. As mentioned, for every task, a machine-learning model can be built to perform the task using data. However, to define and train such models, data availability is a necessity.

Data availability in such applications is often a major obstacle in defining the models. Performance of the first type of task, the detection of existence of damage, may be feasible with simple monitoring of a structure. The problem often reduces to \textit{novelty detection} \cite{Farrar}. A data-driven model in this case is needed to identify a range of normal-condition characteristics of a structure (e.g.\ its natural frequencies). Subsequently, the structure is monitored and, if its characteristics are observed outside the normal-condition range, an inspection procedure can be triggered. 

For the rest of the tasks, the acquisition of data becomes more difficult. Although a damage-classifying machine-learning model can be defined \cite{Farrar}, an appropriate dataset may rarely be available. Moreover, even performing an artificial-damage and repair procedure as in \cite{Farrar}, the repaired structure may not behave the same way as the original \cite{gardner2021overcoming}. Motivated by the scarcity of data, the field of \textit{population-based structural health monitoring} (PBSHM) has developed \cite{gardner2022population}. The intuition behind such an approach to monitoring of structures is that structures and the damage affecting them, have common underlying physics. The common physics are expected to be true especially in the case of similar structures. Thus, knowledge can be transferred from structures which are extensively monitored or structures for which damage-state data are available to newly-built or data-scarce structures. Such transfer facilitates the definition of data-driven models dealing effectively with more tasks in Rytter's hierarchy.

Methods have been developed for the first three steps of the hierarchy \cite{gardner2022population}, as well as for the evaluation of the similarity of structures. For the last two steps, definition of the extent and prognosis, recent work has been performed to take advantage of data from nominally-identical structures, which have reached their failure point, to make probabilistic predictions about currently-failing structures \cite{tsialiamanis2024meta}. In the aforementioned work, the data of extensively-monitored structures are projected on a functional subspace, which corresponds to the damage-evolution mechanism of the population, and, given partial damage-evolution data from new structures, a sampling procedure is followed to make predictions about the evolution of damage of the new structures.

A major assumption made to perform damage prognosis within a population is that the damage which is evolving in the structures of the population follows the same physics across the population. Furthermore, it is assumed that high-fidelity data are available for every structure, training and testing. However, this assumption may contradict the need for low-cost monitoring systems. Within a population damage may evolve in different ways and this may be reflected in the data which become available to the monitoring system. It might also be infeasible to acquire high-fidelity data regarding the extent of damage for every structure of the population, because of the availability or the cost of a high-fidelity monitoring system. 

The current work discusses the problem of allocating monitoring resources within a population. The allocation is performed based on the ability of an existing model to approximate the behaviour so-far of damage evolving within the population. At the same time, an \textit{active-learning} \cite{hughes2022risk} methodology is proposed, with a view to allocating the monitoring system to the structure which shall offer the most informative data to update the existing model and make more accurate predictions in the future. In absence of a specific metric for how informative data samples are, the assumption is made that curves which exhibit high errors will also be highly informative, because the errors might be caused by physics which are not represented by the currently-available data.

The layout of the paper is as follows. In Section \ref{sec:PB_damage_prognosis}, a brief description of a method for a population-based approach to damage prognosis. In Section \ref{sec:active_learning}, a methodology is presented for the evaluation of the performance of the model and the selection of the structures which shall be included in the updating of the model. In Section \ref{sec:application}, a simple simulated crack-growth dataset is used to evaluate the efficiency of the algorithm. Finally, in Section \ref{sec:conclusions}, conclusions are drawn and future steps and extensions to the current work are discussed.

\section{Population-based damage prognosis}
\label{sec:PB_damage_prognosis}

The evolution of damage within a structure is a quite complicated process, which often has inherent uncertainty. Common types of damage, such as crack-growth in metal plates \cite{corbetta2014real}, corrosion in pipes \cite{heidary2018review}, scour in wind turbines \cite{smith2024anomaly}, evolve following physics which might be difficult to model following a \textit{physics-based} approach, however; because of the common physics within a population, data can be exploited to build population-based models to model their evolution. A data-driven and population-based approach to modelling the evolution of damage is presented in \cite{tsialiamanis2024meta}.

The methodology presented in \cite{tsialiamanis2024meta} considers a \textit{training population} of structures, for which detailed data up to their failure (or up to the point when the structure is considered non-usable) are available. Assuming that the fundamental damage mechanism is common for the population, and the differences in the behaviour can be explained by varying parameters of the mechanism, a functional subspace of the damage-evolution curves is calculated. Considering that the damage curves are observed at times $\mathbf{t}=[t_{1}, t_{2}..., t_{N_{t}}]$ (or equivalently number of loading cycles), data are available in the form,
\begin{equation}
    \label{eq:training_data}
    D = \{ h^{1}, h^{2}..., h^{N} \}, \quad h^{i} = \{(t_{1}, y^{i}_{1}), (t_{2}, y^{i}_{2})..., (t_{N_{t}}, y^{i}_{N_{t}})\}, \quad i=1, 2..., N_{s}
\end{equation}
where $y^{i}_{j}$ is a feature describing the extent of damage (e.g.\ a crack length) at the time $t_{j}$ for the $i$\textsuperscript{th} structure, and $N_{s}$ is the number of structures in the training population. In the above equation, a damage curve $h^{i}$ is a function which describes the extent of damage as a function of time, e.g.\ $h^{i}(t_{j}) = y^{i}_{j}$.

To create a parametrised data-driven model of the available data, the curves $h_{i}$ of equation (\ref{eq:training_data}) are decomposed using \textit{functional principal component analysis} (fPCA) \cite{ramsay2005principal} and the equations of the curves are written as,
\begin{equation}
    \label{eq:fPCA}
    h^{i}(t) = \sum_{j=1}^{j=K} \beta^{i}_{j} \varphi_{j}(t) + \epsilon
\end{equation}
where $\varphi_{j}(t)$ are the basis functions of the fPCA decomposition of the data, $K$ is the number of principal components used in the decomposition, and $\beta^{i}_{j}$ are the coefficients of the basis functions which compose the $i$\textsuperscript{th} curve $h^{i}$. Therefore, the relationship of the evolution of damage is parametrised by the principal component coefficients $\beta_{j}$.

Following \cite{tsialiamanis2024meta}, damage prognosis is performed given partial data from testing structures in the form,
\begin{equation}
    \label{eq:testing_data}
    h^{i} = \{(t_{1}, y^{i}_{1}), (t_{2}, y^{i}_{2})..., (t_{M}, y^{i}_{M})\}, \quad i=1, 2..., N_{st}, \quad M < N_{s}
\end{equation}
where $N_{st}$ is the number of testing structures, and $h^{i}$ is a testing curve which is partially- and biased- observed. In this case, the prognosis reduces to the definition of the principal-component coefficients of equation (\ref{eq:fPCA}) for the new curve, which can be performed using a \textit{Hamiltonian Monte Carlo} (HMC) \cite{girolami2011riemann} search in the space of the coefficients. In \cite{tsialiamanis2024meta}, a further restriction in the subspace is imposed by calculating a prior belief for the HMC algorithm using a \textit{normalising flow} \cite{papamakarios2021normalizing} to describe a quite complicated form of a probability density function in the total space of the coefficients, which is the $\mathbb{R}^{K}$.

The algorithm is able to perform damage prognosis in a probabilistic manner without any knowledge of the underlying physics of the structure or the damage mechanism. It was tested using an experimental crack-growth dataset. It was shown how a part of the available data can be used to calculate the fPCA bases and the prior belief over the total space of the principal component coefficients and how these two could then be used to make predictions for a plate with a growing crack. 

Nevertheless, the algorithm depends on the similarity of the underlying physics and the formation of the prior belief. The two objects which need to be defined (the fPCA basis and the prior belief), affect the potential performance of the model on new testing structures. On the one hand, the basis via its components describes the range of physics which the model may explain. On the other hand, the prior belief describes the potential instances of the specific physics and how probable a damage-evolution scenario is. Moreover, building such a model requires data in the form of equation (\ref{eq:training_data}), which assumes high-fidelity accurate data of the damage-evolution process.

To address these issues and to maximise the performance of the model within the context of a growing population, an active-learning framework is proposed in the current work. Active learning is a strategy of machine-learning which aims at selecting the data which shall be included in the training dataset in order to maximise the performance of the model, or with other goals, such as risk minimisation \cite{hughes2022risk}. Within the current framework, active learning is considered within a framework of limited high-fidelity monitoring resources. The strategy is used in combination with the described damage-prognosis algorithm to decide on which of the structures of a testing population should be monitored in detail and should be included in the updating of the available model to maximise its performance in making predictions for future testing structures.

\section{Active learning for population-based damage prognosis}
\label{sec:active_learning}

Population-based SHM includes long-term monitoring of populations of structures. Within the duration of the monitoring, different damage events might occur to structures in different times. The damage may occur in a similar way and position to the structure, because of the common underlying physics of the structure itself, but also of the environment. However, because the structures are not actually identical, and because of the uncertainties of the environmental conditions, variations in the damage-evolution process may be observed. The previously-described methodology aims at capturing such variations by defining a functional subspace which characterises them. 

As in every machine-learning model application, the testing structures should belong to the same domain as the training. This requirement translates into two conditions which need to be satisfied in order for such a model to be efficient. The first is that the testing structures should be similar to the ones used to build the model and the second is that the type of damage (and its phenotype) should also be similar. If one of the two conditions are not satisfied, the previously-described model may fail to yield accurate estimations of the evolution of the damage. The failure may come from two elements of the model. The first is the functional basis $\Phi = [\varphi_{1}, \varphi_{2}..., \varphi_{K}]$, which describe the underlying functional components of the damage evolution. A type of damage which is not included in the training population may not be sufficiently described by any linear combination of the components of $\Phi$, resulting this way in failure in the predictions. The second element of the model which might cause such a failure is the prior belief regarding the coefficients $\beta_{j}$ of the linear combinations of $\Phi$. This belief is formed by the observed data of the training population and may be false regarding new structures.

As more and more data become available from monitored structures or as new structures enter the population, the updating of the model is needed to resolve the aforementioned problems. A first consideration regarding the updating of the model is that the augmentation of the training population should be performed using high-fidelity data. Approximations or intermediate models may be used to evaluate the timeseries of the extent of damage, which is then used to inform the model to make predictions. Such low-fidelity data may not be appropriate to augment the functional basis or the belief in the coefficients, because of the presence of noise or the possibility that the intermediate model may not be accurate enough. 

To deal with such an issue in time, the current work proposes an active-learning framework. As a strategy, active learning provides indications regarding the data which should be labelled to maximise the efficiency of a model or to minimise its risk. In a similar framework, the current work uses active learning to decide on which structures at a given time-instant should be monitored in detail. The detail of the monitoring refers to a low- or high-fidelity system which may be used to monitor structures within the population, the data from which are used to perform damage prognosis. 

Provided that a restriction in the availability of the high-fidelity system is present, one may have to decide on which structure from the currently-monitored population should this system be assigned to. To make such a decision, the potential effect of the newly-acquired data to the updating of the model should be taken into account. A consideration to make is that a new type of damage or new physics regarding the damage evolution should be included in the training data and the definition of the damage-prognosis model. As a result, the currently-available model must be evaluated based on its performance so-far on the population. Such a framework may simply translate into checking the ability of the model to approximate the so-far available damage-evolution data. Provided that data in the form of equation (\ref{eq:testing_data}) are available, a linear-combination curve of the basis vectors is fitted to the available data and is used to estimate the progress of the damage evolution in the future. From the available data, an error metric can be calculated, which can be used to evaluate potential outliers in the testing population, whether their difference comes from a functional difference in the damage procedure or from the point being in the low-probability areas of the prior belief of the coefficients $\beta_{j}$. The error metric $e^{i}$ of the $i$\textsuperscript{th} structure may be simply defined as the mean-squared error given by,
\begin{equation}
    \label{eq:error_metric}
    e^{i} = \frac{1}{M}\sum_{j=1}^{M}(\hat{h}_{j}^{i} - y^{i}_{j})^{2}
\end{equation}
where $M$ is the number of available damage-evolution samples, $\hat{h}^{i}_{j}$ is the estimated damage-feature value from the application of the prognosis algorithm for the $j$\textsuperscript{th} sample of the $i$\textsuperscript{th} testing structure, and $y^{i}_{j}$ is the $j$\textsuperscript{th} observed damage-feature sample of the $i$\textsuperscript{th} testing structure.

As damage evolves in different testing structures, monitoring $e^{i}$ can indicate high errors regarding the ability of the prognosis algorithm in interpolating the damage evolution so-far of a specific structure. Such an indication is used in the current work as a tool to decide on the structure which should be monitored using the high-fidelity data-acquisition system. The decision shall lead in an extra high-fidelity curve for the updating of the model. The assumption is made that the curve which currently exhibits the highest error will allow for the updating of the model in a way which will reduce the total error over the population. The update comes from including an extra curve to calculate the fPCA basis, as well as to define the prior belief for the HMC curve-fitting. In the following section, a simple simulated example of the described method is presented.

\section{Numerical application}
\label{sec:application}

\begin{figure}
    \centering
    \includegraphics[width=0.49\textwidth]{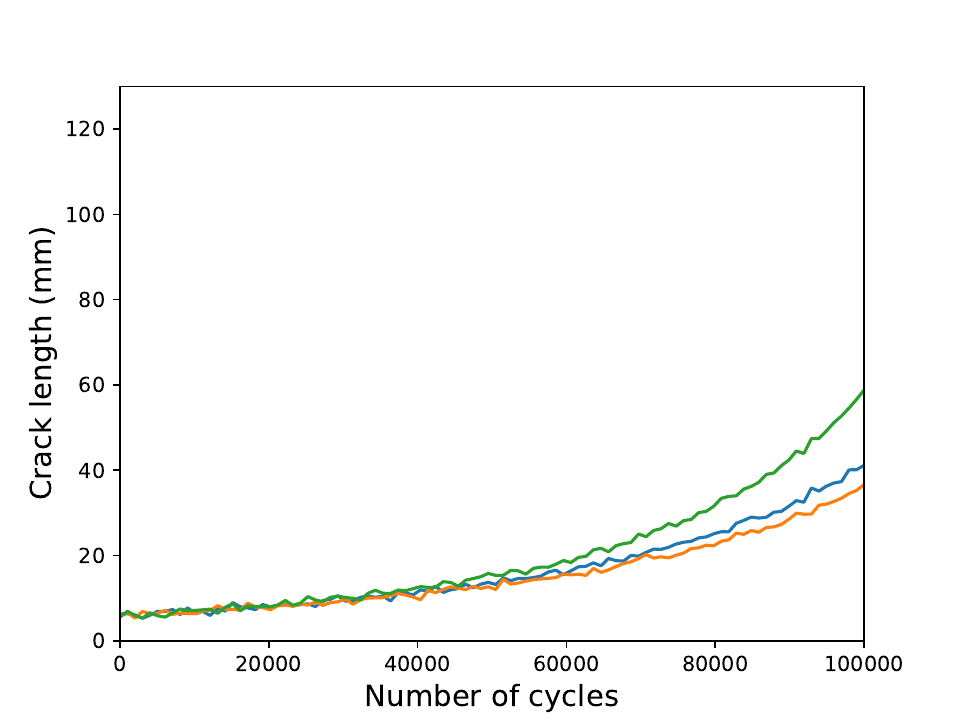}
    \includegraphics[width=0.49\textwidth]{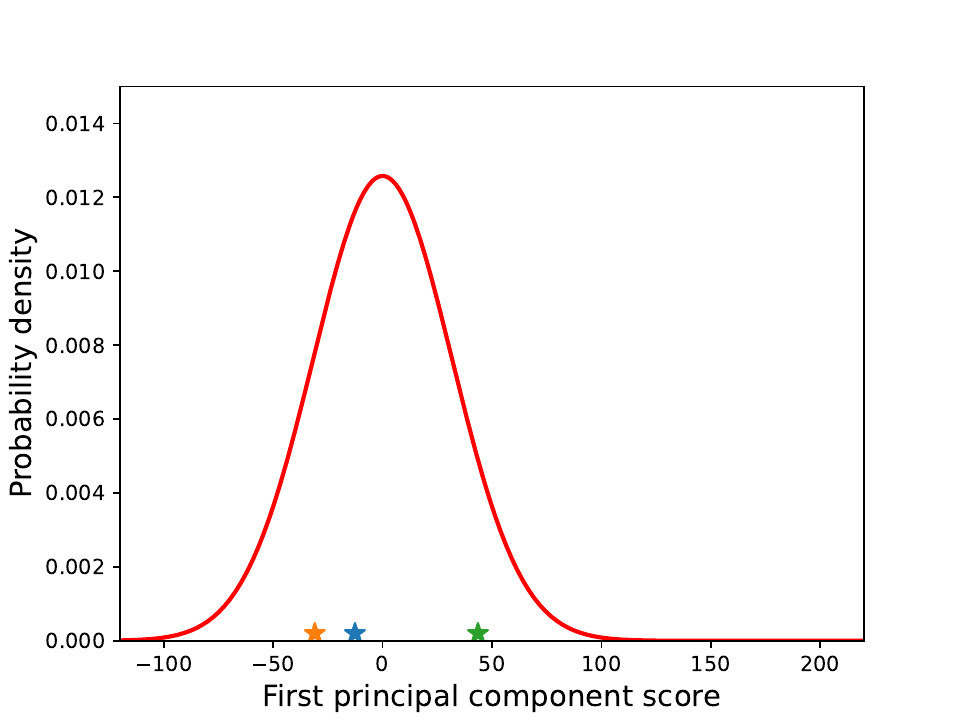}
    \caption{The high-fidelity data of the initial population used to build the first instance of the prognosis model (left), and the corresponding projections on the principal component axis (coloured stars on the right-hand side) and the Gaussian distribution prior defined based on these principal components.}
    \label{fig:initial_pop}
\end{figure}

The described methodology is evaluated using a simulated dataset. The dataset is created using the Paris-Erdogan law \cite{paris1963critical} for crack growth. Initially, a population of three damage curves is created, Figure \ref{fig:initial_pop} on the left. On the right, the values of the $\beta$ for each curve is shown, which corresponds to the principal component score of each curve. In the same figure, the prior belief which is defined simply as a Gaussian distribution over the three samples is shown. It is interesting to note that higher values of $\beta$ correspond to faster-growing curves. The Paris-Erdogan equation parameters used for the curves are given in Table \ref{table:1}. All the crack-growth curves in the current work are discretised in $100$ equidistant samples along the x-axis. Using this population, a functional basis $\Phi$ is created with one principal component explaining over $95\%$ of the variance of the curves. Subsequently, the three curves are projected on the basis and their principal component scores, corresponding to the coefficients $\beta^{i}$, are calculated. Using these three values, the prior belief is defined simply as a Gaussian distribution with mean and standard deviation calculated from the three samples.

\begin{table}[h!]
\begin{center}
\begin{tabular}{ |p{5cm}||p{2cm}|p{2cm}|p{2cm}|  }
 \hline
\backslashbox{Parameter}{Curve} & Blue & Orange & Green \\
 \hline
 Material coefficient $m$   & $2.65$    & $2.65$&   $1.001 \times 2.65$\\
  \hline
 Material coefficient $C$   &   $6\times 10^{-13}$  & $5.58\times 10^{-13}$   & $6.72\times10^{-13}$\\
  \hline
 Stress amplitude (MPa) $\Delta \sigma$ & $300$ & $300$&  $300$\\
  \hline
 Initial crack length $a_{0}$ (mm)   & $3$ & $3$&  $3$\\
 \hline
\end{tabular}
\end{center}
\caption{Training population Paris-Erdogan equation parameters.}
\label{table:1}
\end{table}

\begin{figure}
    \centering
    \includegraphics[width=0.49\textwidth]{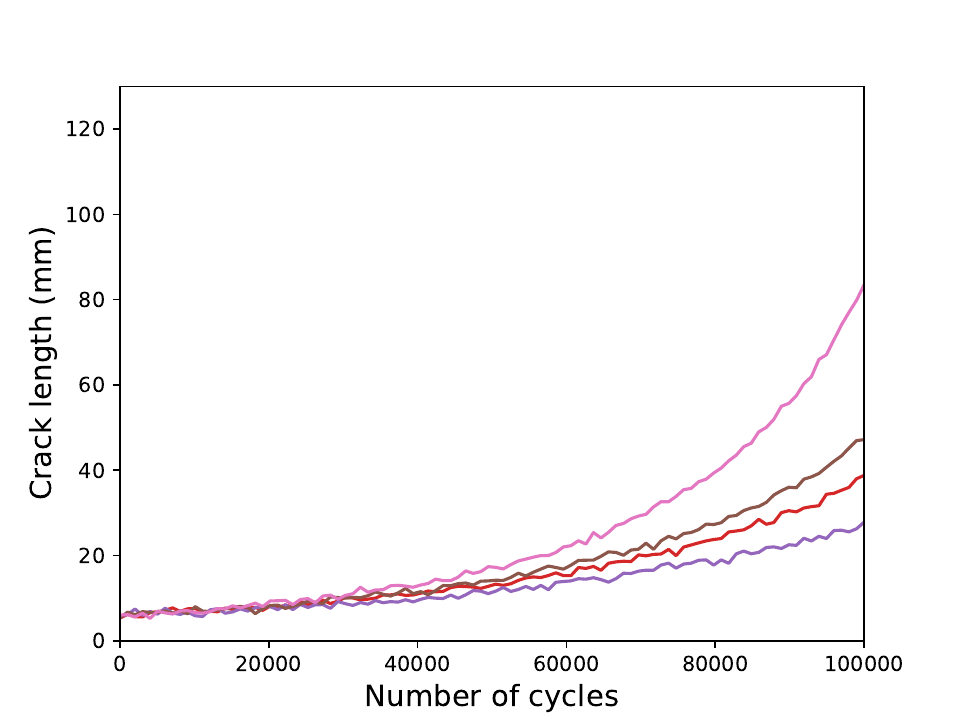}
    \includegraphics[width=0.49\textwidth]{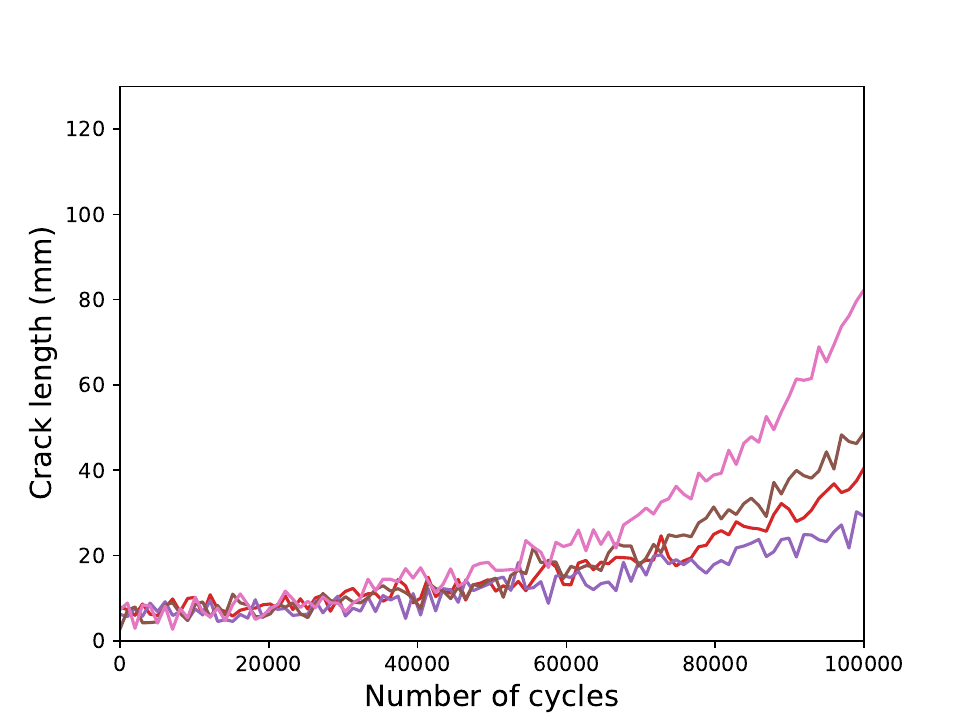}
    \caption{The high-fidelity crack-growth data of the first testing population (left) and the corresponding low-fidelity crack-growth curves (right).}
    \label{fig:first_testing_pop}
\end{figure}

The available model is then used to make predictions for a first testing population, where, for each curve, the error metric of equation (\ref{eq:error_metric}) is monitored. The true high-fidelity crack-growth data of the first testing population is shown in Figure \ref{fig:first_testing_pop} on the left and the corresponding low-fidelity data are shown in the same Figure on the right; the low fidelity data are generated by adding higher level of noise to the underlying crack-growth curves. The corresponding Paris-Erdogan parameters are shown in Table \ref{table:2}. Such low-fidelity data may come from a passive monitoring system, e.g.\ the use of strain fibre to evaluate the crack length, or some visual-based system. The high-fidelity data may come from actual measurements of the crack, which costs more in terms of economical and human resources. 

\begin{table}[h!]
\begin{center}
\begin{tabular}{ |p{5cm}||p{2.2cm}|p{2.2cm}|p{2.2cm}| p{2.2cm}| }
 \hline
\backslashbox{Parameter}{Curve} & Red & Purple & Brown & Pink \\
 \hline
 Material coefficient $m$   & $0.995\times2.65$    & $0.999\times2.65$&   $1.0015 \times 2.65$ & $1.005 \times 2.65$ \\
  \hline
 Material coefficient $C$   &   $6.42\times 10^{-13}$  & $5.1\times 10^{-13}$   & $6.12\times10^{-13}$ & $6.72\times10^{-13}$\\
  \hline
 Stress amplitude (MPa) $\Delta \sigma$ & $300$ & $300$&  $300$ & $300$\\
  \hline
 Initial crack length $a_{0}$ (mm)   & $3$ & $3$&  $3$& $3$\\
 \hline
\end{tabular}
\end{center}
\caption{First testing population Paris-Erdogan equation parameters.}
\label{table:2}
\end{table}

\begin{figure}
    \centering
    \includegraphics[width=\textwidth]{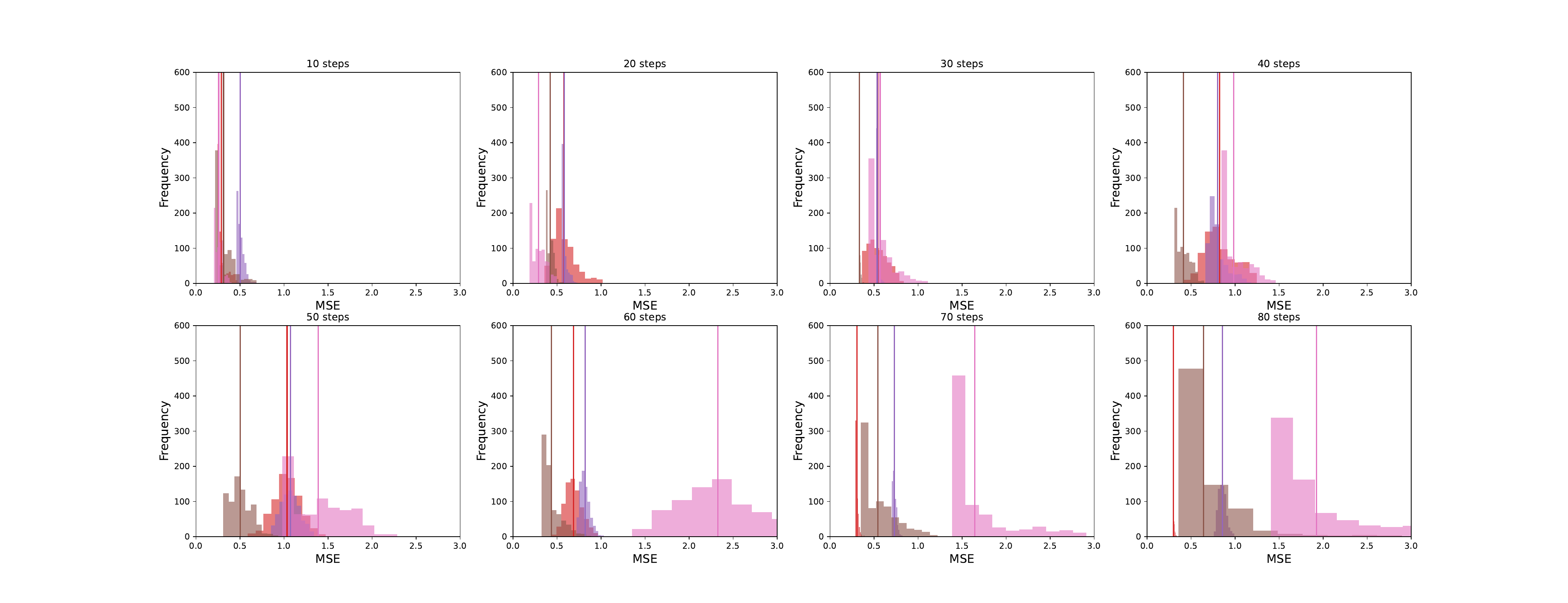}
    \caption{Histograms of the error metrics of equation (\ref{eq:error_metric}) for different number of available samples $M$. The colours correspond to the same-colour curves of Figure \ref{fig:first_testing_pop} and the vertical lines correspond to the mean values of the errors.}
    \label{fig:error_progression}
\end{figure}

Considering that the structures are monitored using data from a part of the curves shown in Figure \ref{fig:first_testing_pop}, the error metric is calculated for every HMC sample for different number of available samples. The total number of samples for each curve is $100$, which is defined by the followed discretisation. The error histograms are presented in Figure \ref{fig:error_progression} together with the corresponding mean value for number of available samples equal to $10, 20, 30..., 80$. The errors of the brown and the red curve are relatively low for the whole procedure, which is to be expected, because they are quite similar to the initial-population curves of Figure \ref{fig:initial_pop}. However, the algorithm yields higher errors for the purple and the pink curve, especially after acquiring $40$ crack-length points. Thus, following the proposed framework, at $40$ steps, the high-fidelity monitoring of the pink curve is decided and the curve is included in the augmented training population. 

A problem regarding the inclusion of the pink curve in the dataset is that up to the point of applying the high-fidelity system to the structure, only the low-fidelity data are available. To deal with this issue, the damage-prognosis algorithm is used to interpolate the unobserved curve up to the $40$th step and for the rest of the damage evolution, the high-fidelity curve is used. This approach is expected to yield a satisfactory approximation of the underlying high-fidelity curve, because the fPCA algorithm captures part of the physics even from the outlier curves and also discards the excess noise of the low-fidelity system -- because the fPCA basis is constructed using the high-fidelity data. 

\begin{figure}
    \centering
    \includegraphics[width=0.49\textwidth]{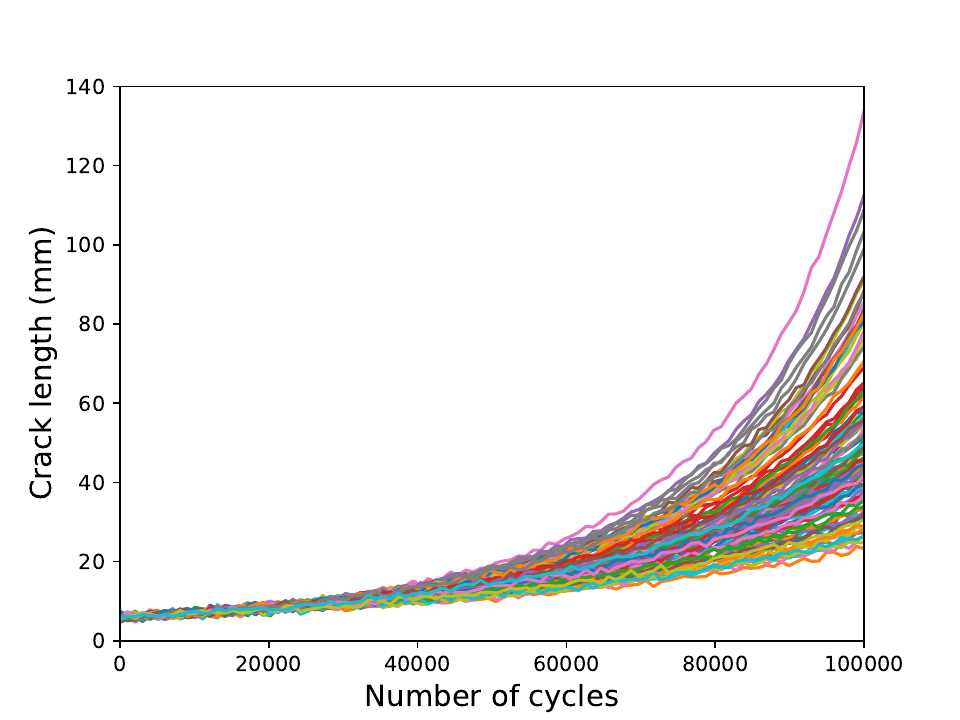}
    \includegraphics[width=0.49\textwidth]{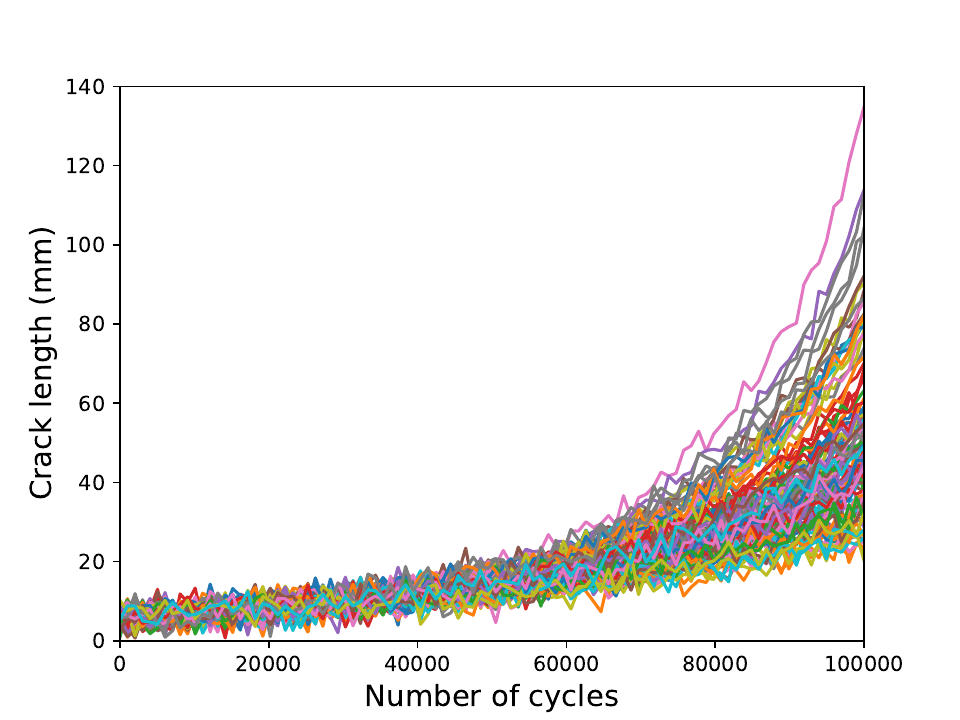}
    \caption{The high-fidelity crack-growth data of the second testing population (left) and the corresponding low-fidelity crack-growth curves (right).}
    \label{fig:second_testing_pop}
\end{figure}

To evaluate the effect of the decision of including the outlier curve in the training dataset, a second testing population is considered. The high- and low-fidelity curves of the new population are shown in Figure \ref{fig:second_testing_pop}.  For the curves in the new testing population, the same prediction procedure is followed using the bases and the priors of the old (not updated) model, and the model updated with the inclusion of each one of the curves in Figure \ref{fig:first_testing_pop} separately. 

\begin{figure}
    \centering
    \includegraphics[width=\textwidth]{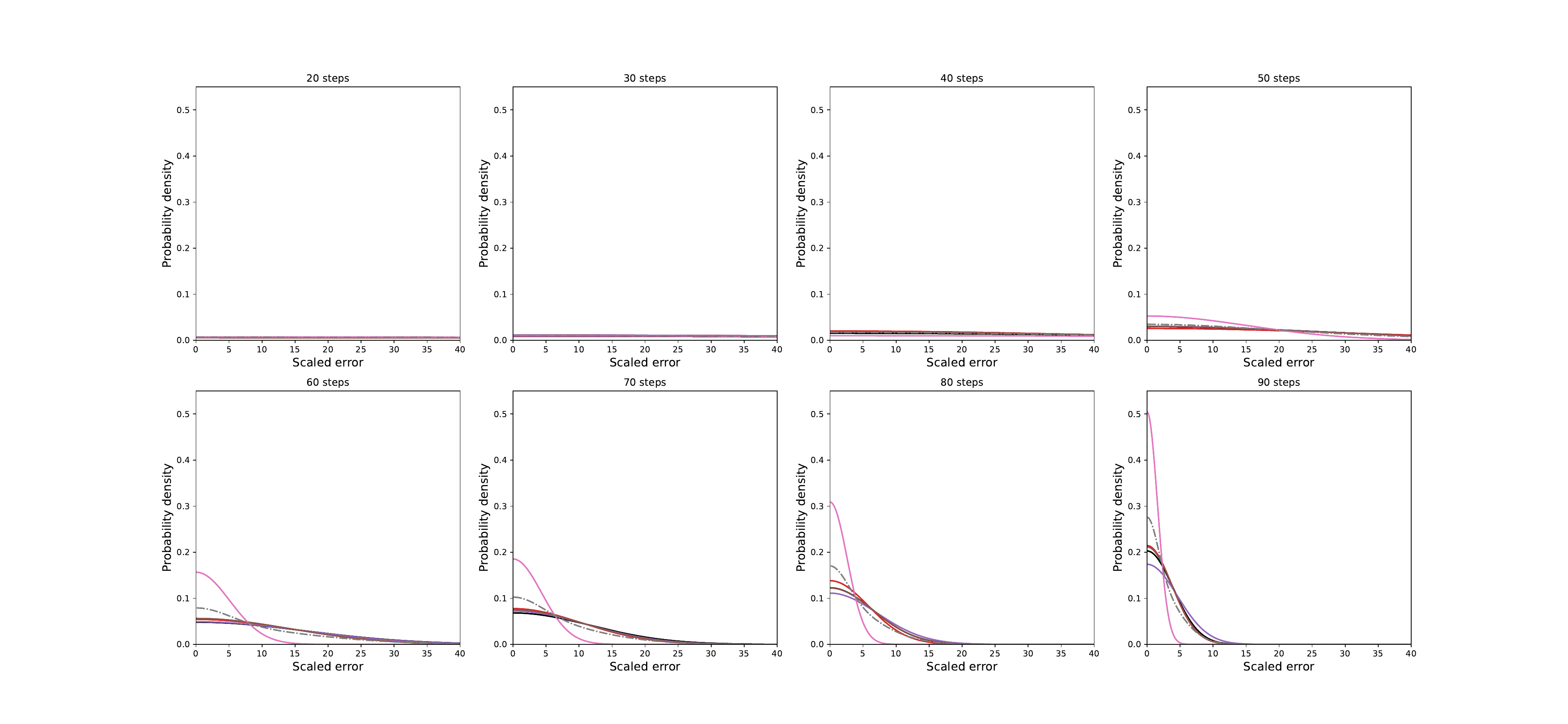}
    \caption{Probability density functions of the scaled errors of the prediction at the highest number of cycles. The black curve corresponds to the errors of the non-updated model. The coloured curves correspond to models updated using the corresponding curve from Figure \ref{fig:first_testing_pop}, the dashed-dotted curve corresponds to the expected error using random structure selection for the updating and the errors are provided for an increasing number of observations.}
    \label{fig:errors_pdfs}
\end{figure}

The probability density of the normalised mean-squared errors on the prediction of the total lifetime of the members of the testing population are shown in Figure \ref{fig:errors_pdfs}. The errors refer to the prediction of the crack length at the maximum number of loading cycles. To account for the risk of structural damage evolving faster than expected, the errors are weighted by a coefficient of $y^{i}_{100} / y_{max}$, where $y^{i}_{100}$ is the maximum crack length of the $i$\textsuperscript{th} curve and $y_{max}$ is the maximum crack length of the whole population. The error is given by $100 \frac{y^{i}_{100}}{y_{max}\sigma^{2}_{y_{100}}}(\hat{y}^{i}_{100}-y^{i}_{100})^{2}$, where $\hat{y}^{i}_{100}$ is the model's estimation of the last-timestep crack length, and $\sigma_{y_{100}}$ is the standard deviation of the maximum crack lengths of the testing population. Such an error is motivated by the normalised mean-squared error, which is used to express the error scaled relative to the spread of the actual values. The coefficient $y^{i}_{100} / y_{max}$ applies more weight to errors corresponding to faster-growing crack-curves. The errors are also presented for different numbers of available samples. 

\begin{figure}
    \centering
    \includegraphics[width=0.24\textwidth]{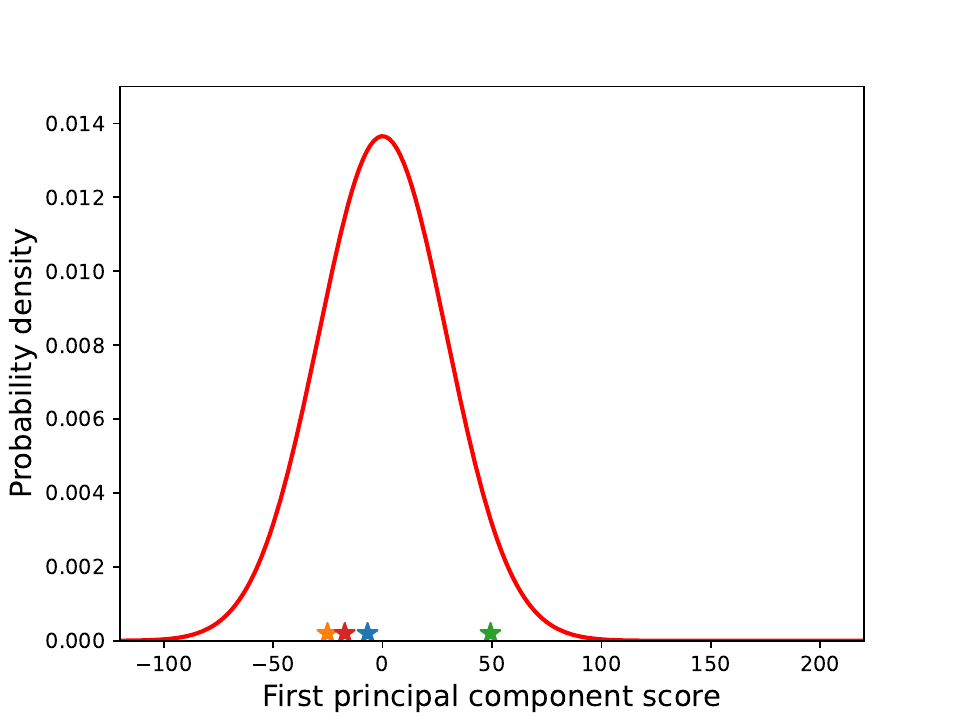}
    \includegraphics[width=0.24\textwidth]{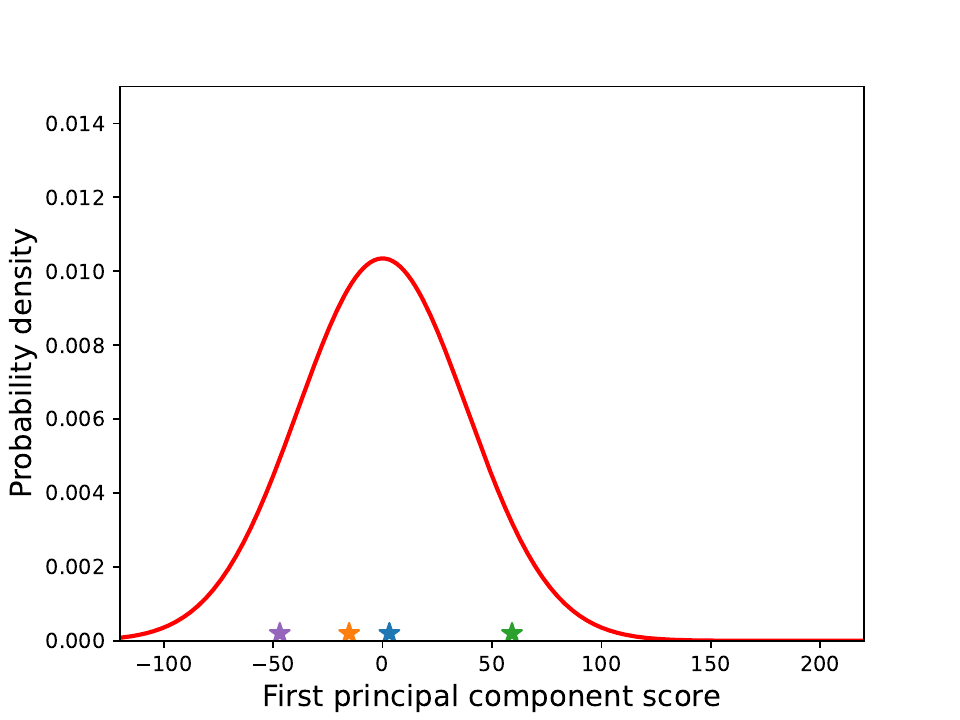}
    \includegraphics[width=0.24\textwidth]{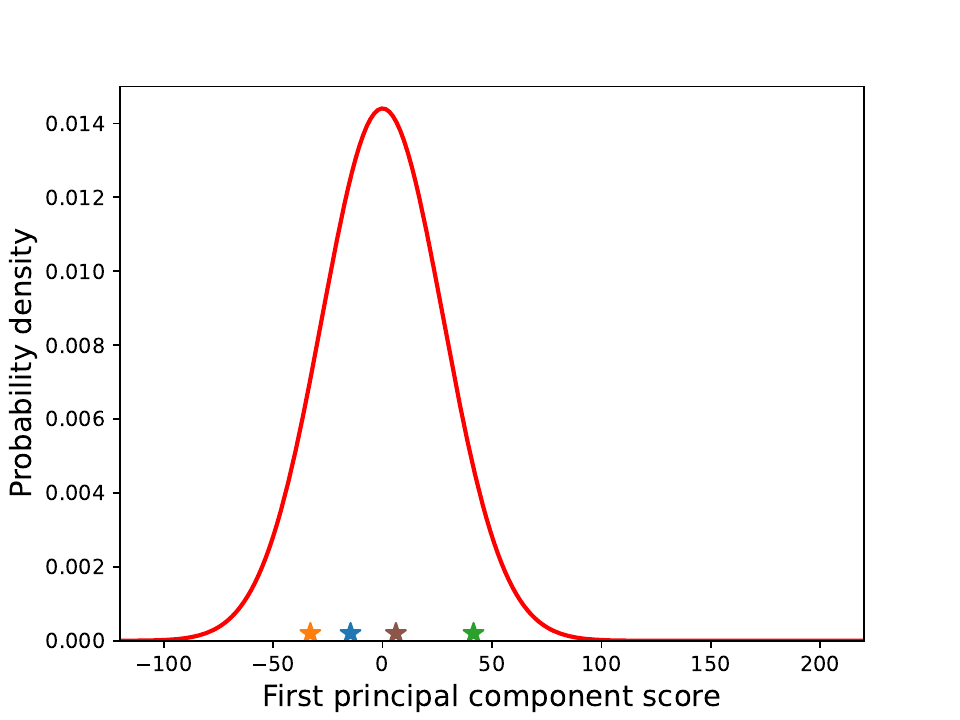}
    \includegraphics[width=0.24\textwidth]{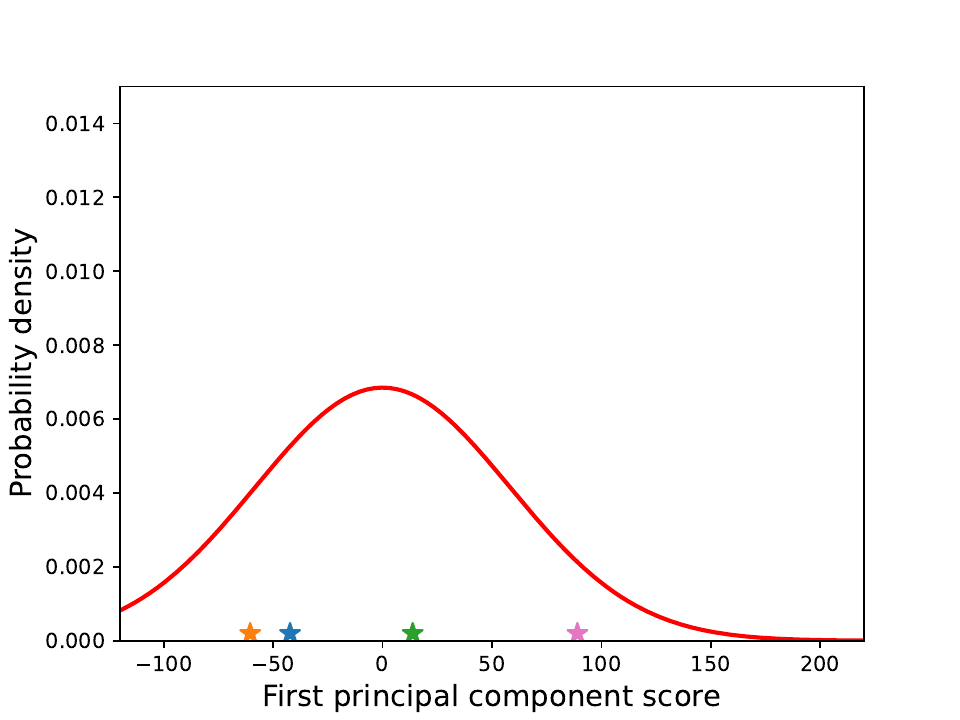}
    \caption{The updated prior beliefs after including a new curve in the training data (red curves). The coloured stars indicate the value of the principal component score of the training curves. The colour of the curves used for the definition of the fPCA basis and the prior belief correspond to the colours of the stars in the same plot.}
    \label{fig:updated_priors}
\end{figure}

As can be seen, for a low number of available samples, the errors are quite spread. However, as more samples become available, the errors start accumulating in lower-value areas. What is particularly interesting, is that the initial model (black curve) performs similarly to the models which are updated using curves other than the pink curve. This similarity may be the result of updating the prior distribution of the $\beta$ coefficient using more samples which are close to each other (the red and brown curves are quite similar to the initial population of the blue, orange and green curves). It may also be seen that in some cases updating with the purple curve results in worse performance. This reduction in performance may be coming from moving the prior belief away from $\beta$ values which correspond to faster-growing curves, which happens by including in the training data the slow-growing purple curve. This belief is confirmed in Figure \ref{fig:updated_priors}, where the updated distributions using a different sample each time are shown. Including any curve other than the pink one results in the prior-belief distribution assigning lower probability values to higher $\beta$ values, which correspond to faster-growing curves. In every case the distribution corresponding to updating the model based on the error criterion of equation (\ref{eq:error_metric}) yields higher density to lower scaled-error values, especially compared to the dashed distribution, which corresponds to the expected error for random selection of the curve to include in the updating, i.e.\ the average of the distributions of the red, purple, brown, and pink distributions.

\section{Conclusions and future steps}
\label{sec:conclusions}

The current work presents an active-learning methodology for damage prognosis and resource allocation within a population of structures. The proposed framework is based on monitoring the performance of a damage-prognosis model on the observed data so-far from degrading structures. An error metric is used to evaluate if a structure is potentially an outlier regarding the predictive abilities of the model. Following an active-learning strategy, the structures which is considered to be the most significant outlier is assigned a high-fidelity monitoring system to acquire the damage-evolution data and update the prognosis model. After the structure has reached a critical limit, its damage-evolution time-history is taken into account to update the existing model with a view to widening its predictive capabilities to more diverse damage-evolution types and structures.

The proposed methodology is evaluated using a crack-growth simulated dataset. An initial training population of three structures is used to define the a first prognosis model. Then, four testing structures, which are monitored using a low-fidelity system, are presented and the ability of the model to interpolate the so-far damage-evolution timeseries is monitored using a simple error metric. Considering limited availability of only one high-fidelity system, the structure which is most-likely an outlier is identified via the aforementioned error metric and the high-fidelity system is assigned to it to acquire its crack-growth curve, and to update the prognosis model. Finally, the performance of the model, which is updated in an informed manner according to the proposed approach, is compared to randomly selecting the structure which would be included in the updating. The results reveal that the informed selection yields lower errors in performing damage prognosis within a new larger testing population of degrading structures.

The proposed algorithm appears to be efficient in selecting the structure which appears to be an outlier. The proposed methodology mainly focussed on the predictive abilities; however, active-learning approaches may incorporate potential risk in the selection of the data. In the current problem, incorporation of risk is a natural thing to do because of the potential problems a quickly-degrading structure may cause. In future work it is expected that the risk of potential failure of a structure will be included in the selection process. An example of how this might affect the current results is that the purple curve also appears to be an outlier, a safer one nonetheless. A structure with quite slowly-evolving damage (purple curve) may not be prioritised compared to one in which damage evolves quicker (pink curve) in case their error metrics are comparable.

The inclusion of physics in the problem might also prove useful. As discussed, in the presented example, higher values of principal components correspond to quicker-growing curves. On the contrary, very low values of principal component scores shall correspond to negative crack-growth curves, which is unnatural. However, a Gaussian distribution like the one defined as a prior in the current work may lead to non-zero probability densities for such principal component scores. As a result, a physics-based definition of the prior belief may be needed to exclude such cases. More sophisticated approaches to the definition of the prior beliefs may also be needed in the case of heterogeneous (either structurally or regarding the evolution of damage) populations. In such cases, multi-modal distributions may be required to characterise different clusters of the population and to allow the current algorithm to search in an area of the parameter space which is informed by the observed populations.

Another issue which might come up when model updating is performed is the potential need to adapt the complexity of the model. An identified outlier might lead to higher complexity model which might have reduced performance in the rest of the population. Thus, future work may need to be focussed on defining local bases and local models during the monitoring and model-updating process. Such approaches can lead to a multi-fidelity-model framework, according to which different models may be more efficient for different members of the population.  

Concluding, the proposed algorithm offers a first step for utilising active learning for resource allocation within a population of structures The performance of the algorithm indicates that such informed resource allocation techniques may prove quite useful for PBSHM and motivates further work in the design of PBSHM systems.


\section*{Acknowledgements}
K.W., A.J.H., and N.D. would like to gratefully acknowledge the support of the UK Engineering and Physical Sciences Research Council (EPSRC) via grant references EP/W005816/1. For the purpose of open access, the authors has applied a Creative Commons Attribution (CC BY) licence to any Author Accepted Manuscript version arising.



\bibliography{References}

\begin{thebibliography}{10}
\providecommand{\url}[1]{#1}
\csname url@samestyle\endcsname
\providecommand{\newblock}{\relax}
\providecommand{\bibinfo}[2]{#2}
\providecommand{\BIBentrySTDinterwordspacing}{\spaceskip=0pt\relax}
\providecommand{\BIBentryALTinterwordstretchfactor}{4}
\providecommand{\BIBentryALTinterwordspacing}{\spaceskip=\fontdimen2\font plus
\BIBentryALTinterwordstretchfactor\fontdimen3\font minus \fontdimen4\font\relax}
\providecommand{\BIBforeignlanguage}[2]{{%
\expandafter\ifx\csname l@#1\endcsname\relax
\typeout{** WARNING: IEEEtran.bst: No hyphenation pattern has been}%
\typeout{** loaded for the language `#1'. Using the pattern for}%
\typeout{** the default language instead.}%
\else
\language=\csname l@#1\endcsname
\fi
#2}}
\providecommand{\BIBdecl}{\relax}
\BIBdecl

\bibitem{Farrar}
C.~Farrar and K.~Worden, \emph{Structural Health Monitoring: A Machine Learning Perspective}.\hskip 1em plus 0.5em minus 0.4em\relax John Wiley and Sons, 2011.

\bibitem{Bishop2}
C.~Bishop, \emph{Pattern Recognition and Machine Learning}.\hskip 1em plus 0.5em minus 0.4em\relax Springer-Verlag, 2006.

\bibitem{goodfellow2016deep}
I.~Goodfellow, Y.~Bengio, and A.~Courville, \emph{Deep learning}.\hskip 1em plus 0.5em minus 0.4em\relax MIT press, 2016.

\bibitem{rytter1993vibrational}
A.~{Rytter}, ``Vibrational based inspection of civil engineering structures,'' Ph.D. dissertation, 1993.

\bibitem{gardner2021overcoming}
P.~Gardner, L.~Bull, N.~Dervilis, and K.~Worden, ``Overcoming the problem of repair in structural health monitoring: {M}etric-informed transfer learning,'' \emph{Journal of {S}ound and {V}ibration}, vol. 510, p. 116245, 2021.

\bibitem{gardner2022population}
P.~Gardner, L.~Bull, J.~Gosliga, N.~Dervilis, E.~Cross, E.~Papatheou, and K.~Worden, ``Population-{B}ased {S}tructural {H}ealth {M}onitoring,'' \emph{Structural {H}ealth {M}onitoring {B}ased on {D}ata {S}cience {T}echniques}, pp. 413--435, 2022.

\bibitem{tsialiamanis2024meta}
G.~Tsialiamanis, C.~Sbarufatti, N.~Dervilis, and K.~Worden, ``On a meta-learning population-based approach to damage prognosis,'' \emph{Mechanical {S}ystems and {S}ignal {P}rocessing}, vol. 209, p. 111119, 2024.

\bibitem{hughes2022risk}
A.~Hughes, L.~Bull, P.~Gardner, R.~Barthorpe, N.~Dervilis, and K.~Worden, ``On risk-based active learning for structural health monitoring,'' \emph{Mechanical {S}ystems and {S}ignal {P}rocessing}, vol. 167, p. 108569, 2022.

\bibitem{corbetta2014real}
M.~Corbetta, C.~Sbarufatti, A.~Manes, and M.~Giglio, ``Real-time prognosis of crack growth evolution using sequential {M}onte {C}arlo methods and statistical model parameters,'' \emph{{IEEE} {T}ransactions on {R}eliability}, vol.~64, no.~2, pp. 736--753, 2014.

\bibitem{heidary2018review}
R.~Heidary, S.~Gabriel, M.~Modarres, K.~Groth, and N.~Vahdati, ``A review of data-driven oil and gas pipeline pitting corrosion growth models applicable for prognostic and health management,'' \emph{International {J}ournal of {P}rognostics and {H}ealth {M}anagement}, vol.~9, no.~1, 2018.

\bibitem{smith2024anomaly}
S.~Smith, A.~Hughes, T.~Dardeno, L.~Bull, N.~Dervilis, and K.~Worden, ``Anomaly detection in offshore wind turbine structures using hierarchical {B}ayesian modelling,'' \emph{arXiv preprint arXiv:2402.19295}, 2024.

\bibitem{ramsay2005principal}
J.~Ramsay and B.~Silverman, ``Principal components analysis for functional data,'' \emph{Functional {D}ata {A}nalysis}, pp. 147--172, 2005.

\bibitem{girolami2011riemann}
M.~Girolami and B.~Calderhead, ``Riemann manifold {L}angevin and {H}amiltonian {M}onte {C}arlo methods,'' \emph{Journal of the {R}oyal {S}tatistical {S}ociety {S}eries {B}: {S}tatistical {M}ethodology}, vol.~73, no.~2, pp. 123--214, 2011.

\bibitem{papamakarios2021normalizing}
G.~Papamakarios, E.~Nalisnick, D.~Rezende, S.~Mohamed, and B.~Lakshminarayanan, ``Normalizing flows for probabilistic modeling and inference,'' \emph{The {J}ournal of {M}achine {L}earning {R}esearch}, vol.~22, no.~1, pp. 2617--2680, 2021.

\bibitem{paris1963critical}
P.~Paris and F.~Erdogan, ``{A critical analysis of crack propagation laws},'' 1963.

\end{thebibliography}


	%
	%
	%



\end{document}